\documentclass{article}
\usepackage{spconf,amsmath,graphicx}
\usepackage{multirow}
\usepackage{xcolor}
\usepackage{subfigure}
\usepackage{array}
\usepackage{graphicx}
\usepackage{hyphenat}
\newcolumntype{L}[1]{>{\raggedright\let\newline\\\arraybackslash\hspace{0pt}}m{#1}}
\newcolumntype{C}[1]{>{\centering\let\newline\\\arraybackslash\hspace{0pt}}m{#1}}
\newcolumntype{R}[1]{>{\raggedleft\let\newline\\\arraybackslash\hspace{0pt}}m{#1}}

\title{U2-Net: A Bayesian U-Net model with epistemic uncertainty feedback for photoreceptor layer segmentation in pathological OCT scans}
\makeatletter
\def\@name{\textit{Jos\'e Ignacio Orlando, Philipp Seeb\"ock, Hrvoje Bogunovi\'c, Sophie Klimscha, Christoph Grechenig}, \\ \textit{Sebastian Waldstein, Bianca S. Gerendas, Ursula Schmidt-Erfurth}\thanks{This work is partially funded by Vienna Science and Technology Fund (WWTF) through project VRG12-009 and a NVIDIA Hardware Grant.}}
\makeatother

\address{\\ Department of Ophthalmology, Medical University of Vienna, Vienna, Austria
}
\begin{document}
%
\maketitle

\begin{abstract}
In this paper, we introduce a Bayesian deep learning based model for segmenting the photoreceptor layer in pathological OCT scans. Our architecture provides accurate segmentations of the photoreceptor layer and produces pixel-wise epistemic uncertainty maps that highlight potential areas of pathologies or segmentation errors.
We empirically evaluated this approach in two sets of pathological OCT scans of patients with age-related macular degeneration, retinal vein oclussion and diabetic macular edema, improving the performance of the baseline U-Net both in terms of the Dice index and the area under the precision/recall curve. We also observed that the uncertainty estimates were inversely correlated with the model performance, underlying its utility for highlighting areas where manual inspection/correction might be needed. 
\end{abstract}

\section{Introduction}
\label{sec:introduction}

\begin{figure}[t]
 \centering
 \includegraphics[width=0.7\columnwidth]{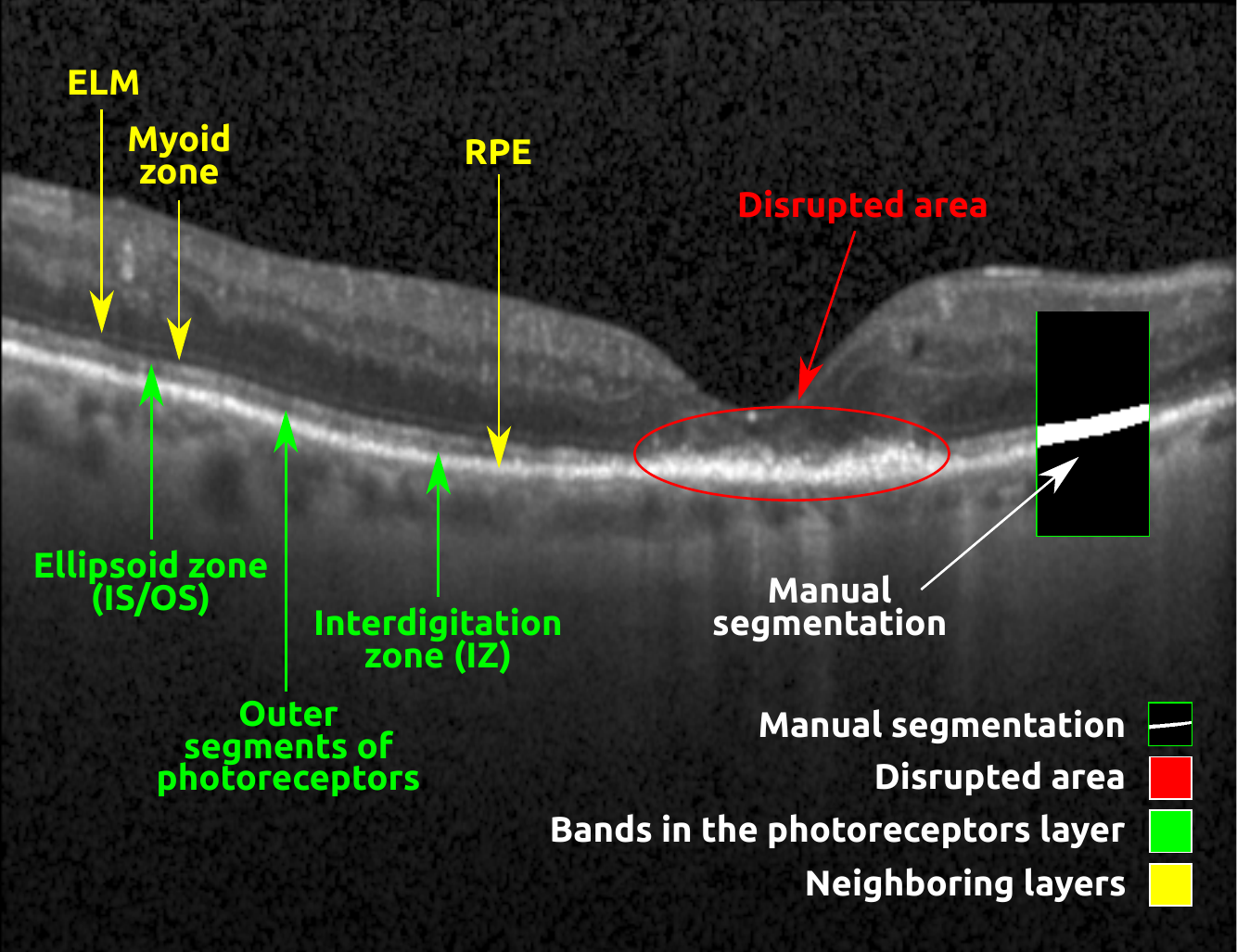}
 \caption{Retinal photoreceptors layer (green) as observed through OCT imaging.}
 \label{fig:layers}
\end{figure}

Age-related Macular Degeneration (AMD)~\cite{wong2014global}, Retinal Vein Occlusion (RVO)~\cite{jonas2010retinal} and Diabetic Macular Edema (DME)~\cite{TAN2017143} are among the leading causes of visual impairment in the world. One common side effect of these diseases is the photoreceptor cell death due to e.g. ischemia or RPE death and neurodegeneration. 
This can eventually affect the visual acuity and/or lead to blindness~\cite{maheshwary2010association}.
Optical coherence tomography (OCT) is a 3D state-of-the-art imaging modality that is currently extensively used for clinical diagnosis and treatment planning in ophthalmology. OCT allows to visually assess the integrity of the photoreceptors, which appear as a layered structure with hyperreflective and hyporeflective bands, located in between the outer limit of the myoid zone and the inner interface of the retinal pigment epithelium (RPE)~\cite{staurenghi2014proposed} (Fig.~\ref{fig:layers}). 
Several clinical applications are benefited by the quantification of the photoreceptor layer characteristics~\cite{ota2008foveal,mori2016restoration}. 
In general, this analysis requires to manually delineate the area for each 2D slice (or B-scan) of the OCT volume, a task that is tedious, time-consuming and prone to a high intra- and intervariability. 
In particular, imaging artifacts such as vessel shadows can be usually labelled as disruptions, 
whilst abnormalities in the photoreceptor thickness lead to diffuse edges that can be difficult to delineate consistently. 
One common way to automatize the segmentation process is to use a supervised learning model. This requires to train a neural network from a large-scale image set that must be representative enough to describe the target population. These sets are hard to acquire for applications such as segmenting the photoreceptor layer in pathological OCTs, as they must cover a large variety of lesions and artifact appeareances. At the same time, regularization techniques such as weight decay or dropout might not be enough to generalize the network to every possible scenario. This drawback can be addressed by also producing uncertainty estimates, as they allow to interpret the outputs and to identify areas where the trained models are ambiguos or not confident about their decision. 

The first automated method for photoreceptor layer segmentation~\cite{chan2006quantification} was based on image processing techniques and evaluated in healthy subjects. More recently, Loo \textit{et al.}~\cite{loo2018deep2} proposed a deep learning strategy to classify the columns of each B-scan (A-scans) as having either healthy or death photoreceptors in OCT images of patients with telangiectasia type 2. 
%
%
In this paper we present the first deep learning approach for segmenting the photoreceptor layer in OCT volumes of patients with AMD, DME and RVO, while also providing qualitative feedback regarding the uncertainty of the model. Our method can be applied to a wider range of diseases than the existing ones~\cite{chan2006quantification}, and its associated pixel-wise uncertainties can be used to recognize pathological areas or regions that need to be corrected.
To this end, we modified the standard U-Net~\cite{ronneberger2015u} architecture (Section~\ref{subsec:network-architectures}) to perform Bayesian inference through Monte Carlo (MC) sampling using dropout at test time. The segmentations were obtained by averaging the MC samples in a pixel-wise way, and the corresponding standard deviation was used to retrieve uncertainty maps~\cite{kendall2015bayesian}. A similar approach was followed in~\cite{nair2018exploring,sedai2018joint}, although combining both epistemic and aleatoric uncertainties. However, we have observed that training our model to predict the aleatoric uncertainty produce worst segmentation results (Section~\ref{sec:results}). 
We performed an extensive evaluation of our model using two data sets of OCT scans with AMD, DME and RVO. We observed that: (i) our modifications of the U-Net combined with MC sampling allows to obtain better results compared to the standard model; (ii) the uncertainty estimates are inversely correlated with the segmentation performance, meaning that this qualitative feedback could be useful to correct the segmentations; and (iii) our method generalizes to disease stages differing from those present in the training set.

\section{Methods}
\label{sec:methods}

\subsection{Bayesian deep learning with dropout}

%
%

Bayesian deep learning allows to compute epistemic uncertainties by modeling a posterior distribution $p(\mathbf{W} | \mathbf{X}, \mathbf{Y})$ over the weights $\mathbf{W}$ of the network, for a given training set of images $\mathbf{X} = \{ X_k \}$ and labels $ \mathbf{Y} = \{ Y_k \}$, $k = 1, ..., N$. In practice, finding the exact posterior is intractable, but an approximation $q(\mathbf{W})$ can be obtained using variational inference, by minimizing the Kullback-Leibler (KL) divergence~\cite{graves2011practical, kendall2015bayesian} $\text{KL}(q(\mathbf{W}) || p(\mathbf{W} | X, Y))$.
In~\cite{kendall2015bayesian}, the authors proposed to model the variational distribution $q(\mathbf{W}_i)$ for the $i$-th convolutional layer of a neural network by using dropout with a $p_i$ dropout-rate. This allows to simultaneously optimize the weights to prevent overfitting while modelling the weights distribution. Dropout can be then used at test time to retrieve multiple Monte Carlo (MC) samples by processing the input $\mathbf{X}$, $T$ times. The resulting outputs can then be averaged to recover a single estimate of the segmentation, and the standard deviation between samples can be taken as an estimate of the epistemic uncertainty.

\subsection{Network architecture}
\label{subsec:network-architectures}

We proposed an \textbf{U}ncertainty \textbf{U}-Net, the U2-Net, that is based on a modified version of the standard U-Net~\cite{ronneberger2015u}. 
The architecture comprised an encoder and a decoder, connected with skip connections. The encoder consisted of 5 convolutional blocks (with 64, 128, 256, 512 and 1024 output channels each) followed by $2 \times 2$ max-pooling. These blocks used two $3 \times 3$ convolutional layers, each of them followed by batch normalization and leaky ReLUs~\cite{xu2015empirical}.
The decoder consisted of 4 upsampling blocks with 512, 256, 128 and 64 output channels. These blocks were modified to use nearest neighbor upsampling followed by a convolutional block. This change allowed to reduce the blocking artifacts that are typical from the transposed convolutions. A final $1 \times 1$ convolutional layer was used at the end to retrieve the final probabilities for background/foreground. To enable epistemic uncertainty estimation, we incorporated dropout after each convolutional block, except for the first and the last blocks. A dropout rate of $p=0.1$ was used in all the cases except for the bottleneck layer, where $p=0.5$ was applied.


\section{Experimental setup}
\label{sec:experimental-setup}

\subsection{Materials}
\label{subsec:materials}

A set of 50 Spectralis OCT scans of 50 different patients (16 DME, 24 RVO, 10 with early AMD and choroidal neovascularization, CNV) was extracted from the study database of the Vienna Reading Center and used in our experiments. Each volume comprised 49 B-scans with 512 columns (A-scans) of 496 pixels each, covering an approximate retinal area of $6 \times 6$-mm, centered on the fovea. Each scan was manually annotated by trained readers, and the resulting labels were manually corrected by experienced ophthalmologists.
This data set was randomly divided into training and validation sets, each of them comprising 31 and 4 volumes. The remaining 15 volumes were used to construct a first test set $A$. A similar proportion of each disease was preserved in the three sets to avoid any bias. Additionally, 10 unseen late AMD volumes with geographic atrophy (GA) were taken to construct the test set $B$. We used the latter to estimate the generalization performance of the method to a more advanced disease stage that was not part of the study data set mentioned before.

\subsection{Baselines and training configuration}
\label{subsec:training-configuration}

We experimentally compared the performance of our U2-Net with three baselines, namely the standard U-Net~\cite{ronneberger2015u}, the BRU-Net~\cite{apostolopoulos2017pathological} (Branch Residual U-Net) and a BU-Net~\cite{nair2018exploring} (Bayesian U-Net). 
Standard graph-based methods~\cite{garvin2009automated} for retinal layer segmentation in OCT scans were not considered as they assume a fixed layer topology, which is inconsistent with the presence of disruptions in the photoreceptor layer.

For the basic U-Net, we incorporated batch normalization and nearest neighbor upsampling to ensure a fair comparison. A dropout rate of $p=0.5$ was set to the bottleneck layer as in~\cite{ronneberger2015u} to increase the regularization.
%
The BRU-Net is a more complex architecture with dilated residual blocks~\cite{apostolopoulos2017pathological}. We used 5 convolutional/upsampling blocks with 32, 64, 128, 256, 512 and 512 output channels respectively due to GPU memory limitations (notice that each block in this baseline architecture is composed of 5 convolutional layers~\cite{apostolopoulos2017pathological}).
%
The BU-Net has two output channels that predict both the segmentation scores and the pixel-wise aleatoric uncertainty estimate $\hat{V}$, which is modelled by adding Gaussian noise with 0 mean and $\hat{V}$ variance to the networks outputs, before applying the softmax activation. Instead of using its original 3D definition~\cite{nair2018exploring}, we used our U2-Net as the core architecture. Thus, the BU-Net in our comparison is not other than our U2-Net but adapted to also learn the aleatoric uncertainty during training.
All the networks were trained at a B-scan level to minimize the cross entropy loss, using Adam optimization and weight decay ($5 \times 10^{-4}$). We used an initial learning rate of $\eta=10^{-4}$ and a batch size of 2 B-scans, and the optimization was performed for a maximum number of 160 epochs. The learning rate was gradually reduced by a factor of 0.5 when the absolute improvement of the average validation Dice index was smaller than $10^{-4}$ during the last 15 epochs. The best model according to the validation Dice was used for evaluation. In all the cases, the segmentations were obtained from the photoreceptor class probability by applying the Otsu thresholding algorithm~\cite{otsu1979threshold}.

\subsection{Evaluation metrics}
\label{subsec:evaluation-metrics}

As our region of interest represents only a small portion of the input B-scan, the segmentation results were evaluated in terms of the area under the precision/recall curve (AUC) and the Dice index, which are not affected by class imbalance~\cite{davis2006relationship}. Alternatively, we also evaluated the ability of the output score maps to deal with disrupted areas of the photoreceptor layer. To this end, we took the maximum photoreceptor probability $y_D$ at each A-scan for a given output, and $1 - y_D$ was taken as a score for layer interruptions. Then, the AUC at an A-scan level was used as an estimate of the model performance for disruption detection.



\section{Results and discussion}
\label{sec:results}

\subsection{Photoreceptor layer segmentation}

\begin{figure}[t]
 \centering
 \subfigure[Photoreceptors]{\includegraphics[width=0.45\columnwidth]{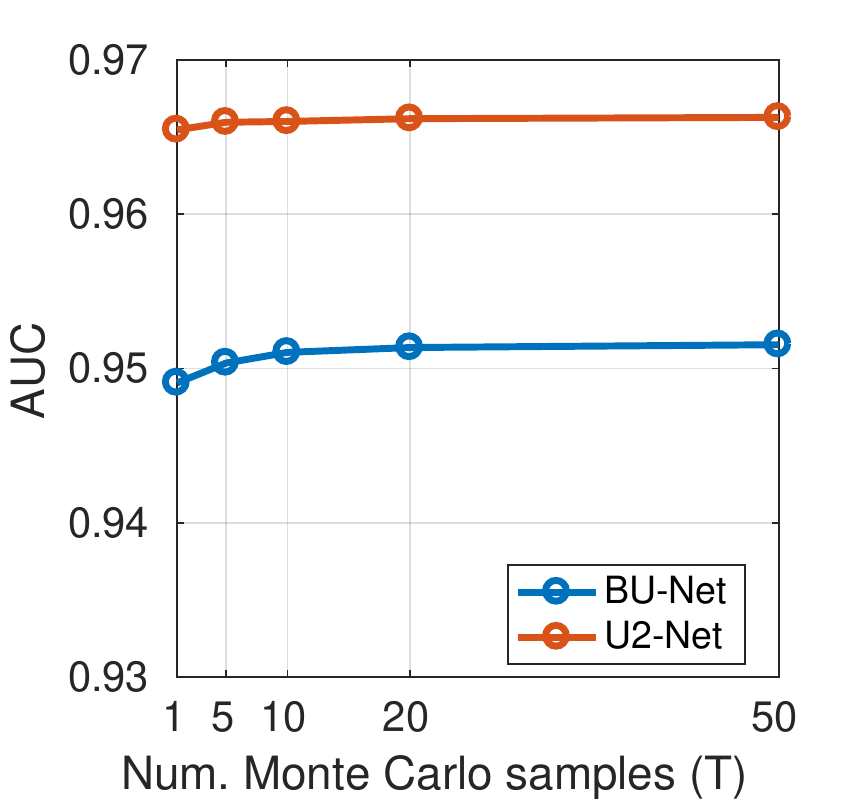}}
 \subfigure[Disruptions]{\includegraphics[width=0.45\columnwidth]{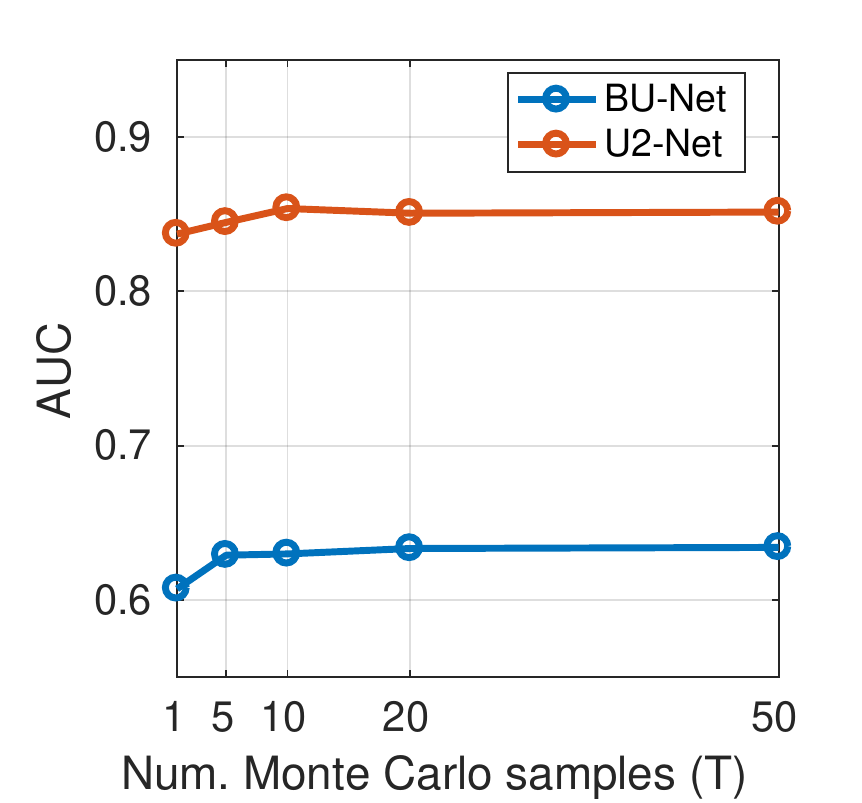}}
 \caption{AUCs in the validation set for different $T$ values.}
 \label{fig:val-t-num}
\end{figure}

\begin{table}[t!]  
\centering
\resizebox{\columnwidth}{!}{
\begin{tabular}{C{1.6cm}||C{1.2cm}|C{1.2cm}|C{1.2cm}||C{1cm}|C{1cm}|C{1.2cm}}
  \hline
  \multirow{4}{*}{\textbf{Model}} & \multicolumn{3}{c||}{\textbf{Test set A}}& \multicolumn{3}{c}{\textbf{Test set B}} \\
  & \multicolumn{3}{c||}{AMD (early, CNV), DME, RVO} & \multicolumn{3}{c}{Late AMD (GA)} \\
  \cline{2-7}
  & \multicolumn{2}{c|}{\textbf{Photoreceptors}} & \textbf{Disruptions} & \multicolumn{2}{c|}{\textbf{Photoreceptors}} & \textbf{Disruptions} \\
  \cline{2-7}
   & \textbf{AUC} & \textbf{Dice} & \textbf{AUC} & \textbf{AUC} & \textbf{Dice} & \textbf{AUC}  \\
  \hline
  \hline
  \textbf{U-Net}~\cite{ronneberger2015u} & 0.9566 & 0.8815 $\pm$0.06 & 0.5077 & 0.9390 & 0.8375 $\pm$0.07 & 0.8795 \\
  \hline
  \textbf{BRU-Net}~\cite{apostolopoulos2017pathological} & 0.9593 & 0.8767 $\pm$0.08 & 0.2621 & 0.9295 & 0.7890 $\pm$0.13 & 0.8333 \\
  \hline
  \textbf{BU-Net} $T=1$                  & 0.9466 & 0.8647 $\pm$0.08 & 0.2222 & 0.8969 & 0.7311 $\pm$0.14 & 0.8065\\
  \hline
  \textbf{BU-Net} $T=10$                 & 0.9505 &  0.8678 $\pm$0.08 & 0.2405 & 0.8998 & 0.7428 $\pm$0.14 & 0.8129\\
  \hline
  \hline
  \textbf{U2-Net} $T=1$                  & 0.9653  & 0.8932 $\pm$0.04 & \textbf{0.6712} & \textbf{0.9500} & \textbf{0.8546 $\pm$0.06} & 0.9085 \\
  \hline
  \textbf{U2-Net} $T=10$                 & \textbf{0.9669} & \textbf{0.8943 $\pm$0.04} & 0.6417 & 0.9472 & 0.8457 $\pm$0.08 & \textbf{0.9101} \\
  \hline
\end{tabular}
}
\caption{Quantitative evaluation on the test sets A and B.}
\label{table:comparison-table} 
\end{table}

We studied the changes in the AUC values when varying the number of MC samples $T$ in the validation set (Fig.~\ref{fig:val-t-num}). It can be seen that the segmentation performance was not significantly improved after $T=20$, while a small drop in the AUC of the disruptions is observed after $T=10$. 

Table~\ref{table:comparison-table} compares the results obtained on the test sets A and B using different segmentation models. The U2-Net achieved the highest performance in the two sets, both for photoreceptor segmentation and disruption detection. In the test set A, an improvement in the segmentation results of the U2-Net was observed when averaging through multiple MC samples, while the performance for disruption detection was slightly decreased. The opposite case was observed in the test set B, where MC sampling improved the performance for disruption detection but affecting the results for photoreceptor segmentation. In all the cases, the BU-Net performed poorly compared both to our U2-Net and the standard network.

\begin{figure}[t]
 \centering
 \subfigure[Dice$=0.9196$, $\overline{u}=6.7\times10^{-4}$]{\includegraphics[width=0.48\columnwidth]{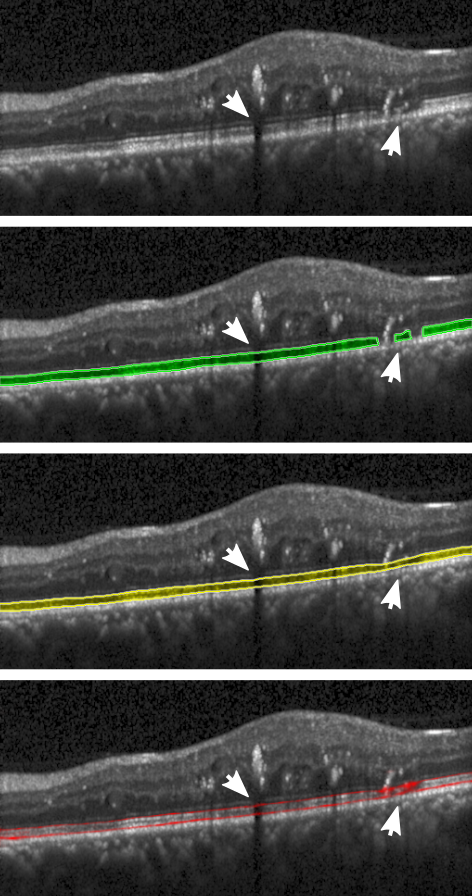}\label{fig:qualitative-results-a}}
 \subfigure[Dice$=0.5888$, $\overline{u}=13\times10^{-4}$]{\includegraphics[width=0.48\columnwidth]{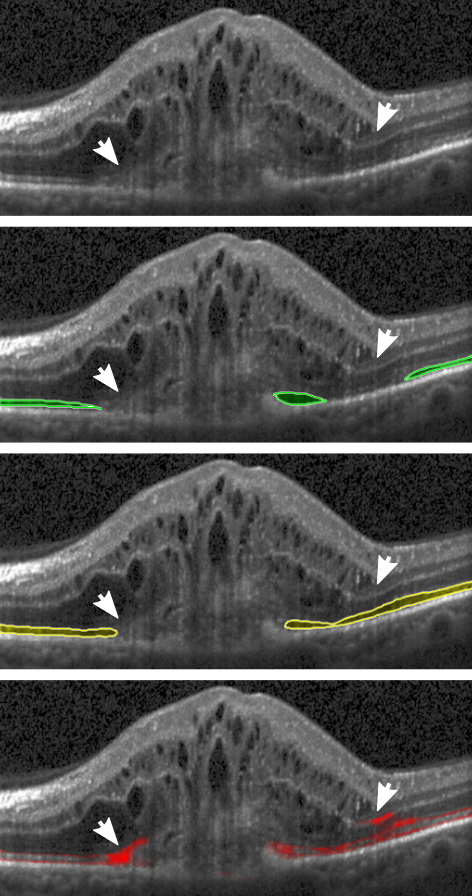}\label{fig:qualitative-results-b}}
 \caption{Qualitative results of the U2-Net ($T=10$) on the test set $A$. From top to bottom: B-scan, manual annotation, automated segmentation, and epistemic uncertainty map. B-scan level Dice and mean uncertainty $\overline{u}$ are also included.}
 \label{fig:qualitative-results}
\end{figure}

Qualitative results of the U2-Net with $T=10$ samples are presented in Fig.~\ref{fig:qualitative-results}, jointly with their associated pixel-wise uncertainty estimates. 
The uncertainty maps were normalized using their maximum value for visualization purposes.

\subsection{Uncertainty estimation}

Fig.~\ref{fig:uncertainty-volume} shows the correlation between the photoreceptor layer segmentation performance (as measured using the Dice index) and the mean uncertainty for each of the volumes on the test set A. The linear regression line is included in the plot to illustrate the general trend of the results. The mean uncertainty at a volume level was observed to be inversely correlated with the segmentation performance ($R^2=0.7644$). A similar behavior was observed at a B-scan level (Fig.~\ref{fig:qualitative-results}).


\subsection{Discussion}

The proposed U2-Net allowed to improve both the segmentation and disruption detection results of the baseline U-Net, outperforming the more complex BRU-Net architecture (Table~\ref{table:comparison-table}). This is a result of a better generalization ability thanks to the incorporation of more dropout than in the baseline networks, and to the ability of leaky ReLUs to prevent vanishing gradients~\cite{xu2015empirical}. 
It was also observed that learning to predict the aleatoric uncertainty of the model (BU-Net) reduced the performance of the architecture in our two test sets (Table~\ref{table:comparison-table}). This might be due to the fact that this additional task acts as a strong regularizer that enforces the network to ignore infrequently occuring 
ambiguities. A general drop in segmentation performance was observed when applying the models on the test set B, although our U2-Net still achieved the best results. The improvement in the AUC values for disruption detection could be explained by the presence of more evident interruptions than those 
in the test set A. 
When averaging multiple MC samples, a trade-off between segmentation and disruption detection performance was observed. This might be caused by the methods producing more (less) true positive responses of the photoreceptors but at the same time more (less) false positive responses in the disrupted areas. Such a behavior is consistent with e.g. being more accurate in terms of the layer thickness or to better segment the photoreceptor layer under vessel shadows, but at the cost of ignoring small disruptions (Fig.~\ref{fig:qualitative-results-a}). Nevertheless, the MC sampling procedure has the added value of providing pixel-wise uncertainties that can be used to correct these errors. It is worth mentioning also that the dropout parameters of the U2-Net were fixed without fine tuning on the validation set. Further improvements in the results could be achieved by finding an optimal configuration.

\begin{figure}[t]
 \centering
 \includegraphics[width=0.7\columnwidth]{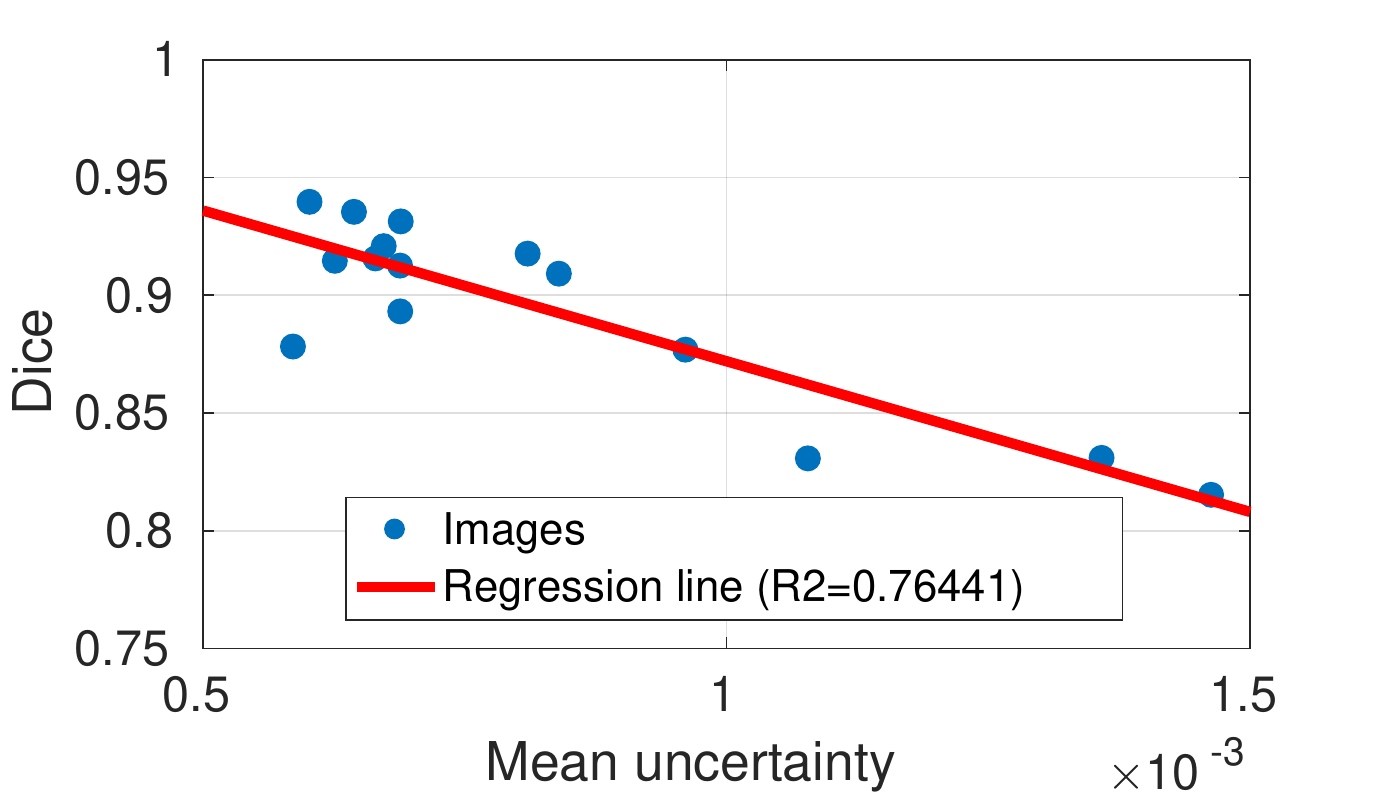}
 \caption{Correlation between segmentation performance and mean uncertainty for each OCT volume in the test set $A$.}
 \label{fig:uncertainty-volume}
\end{figure}

In healthy photoreceptor layers, most of the epistemic uncertainty occurs in the upper and lower interfaces of the segmentation (Fig.\ref{fig:qualitative-results-a}, left arrow). This is in line with the high inter-observer variability observed during the manual annotation process. Under challenging scenarios with small disruptions (Fig.\ref{fig:qualitative-results-a}, right arrow) the model sometimes missinterpreted the area as a layer thinning, although with high uncertainty.
In extremelly pathological scenarios 
(Fig.\ref{fig:qualitative-results-b}), the U2-Net was able to identify most of the largest areas of cell death in the layer, which are more evident due to the concomitant appeareance of cysts and retina thickenning. In Fig.\ref{fig:qualitative-results-b}, most of the wrong predictions were associated to areas of subtle disruptions, differences in the layer thickness (right arrows), or to inconsistencies in the edges of the disruptions (left arrow). In any case, high uncertainty values were observed in those areas. This indicates that the errors in the segmentation could be pointed out by the uncertainty estimates, and subsequently be manually corrected by human readers. This claim is also supported by the correlation analysis in Fig.~\ref{fig:uncertainty-volume}.

\section{Conclusions}
\label{sec:conclusions}

In this paper we presented U2-Net, a deep learning architecture for photoreceptor layer segmentation in pathological OCT scans. Apart from being more accurate than other baseline approaches--including more complex networks~\cite{apostolopoulos2017pathological}--our method provides epistemic uncertainty maps that could be used jointly with the segmentations to accelerate the manual labelling process. We have observed that the mean epistemic uncertainty was inversely correlated with the segmentation performance, meaning that this qualitative feedback could be used by image graders during the manual annotation process to assess the quality of the results and to identify potential errors or pathological cases. Our experiments on a separate test set show that the model was robust enough to deal with different stages of the diseases that were not included on the training set. Hence, by incorporating uncertainty estimations and a modified U-Net architecture, we both improve the generalizability and the interpretability of our model during test time, favoring its application in clinical scenarios~\cite{schmidt2018artificial}. Further research will be performed to improve the results in areas of high uncertainty, and to correlate the uncertainty outcomes with disagreements between human observers.

\bibliographystyle{IEEEbib}
\bibliography{references}

\end{document}